\documentclass[doublecol]{epl2} 
\usepackage{amssymb}
\usepackage{url}

\title{Machine learning in physics: a short guide}
\shorttitle{Machine learning in physics: a short review} 
\author{Francisco A. Rodrigues}

\institute{                    
  \inst{1} Instituto de Ciências Matemáticas e de Computação, Universidade de São Paulo, São, Carlos, São Paulo, Brazil\\
}

\abstract{
Machine learning is a rapidly growing field with the potential to revolutionize many areas of science, including physics. This review provides a brief overview of machine learning in physics, covering the main concepts of supervised, unsupervised, and reinforcement learning, as well as more specialized topics such as causal inference, symbolic regression, and deep learning. We present some of the principal applications of machine learning in physics and discuss the associated challenges and perspectives.
}
\begin{document}

\maketitle

Ernest Rutherford once declared: ``if your experiment needs statistics, you ought to have done a better experiment''~\cite{Vaux}. His remark reflects his belief in the significance of well-controlled experiments and the need for experimental designs that minimize uncertainties and sources of errors. However, while Rutherford's statement may have merit in his time, it no longer applies in the modern scientific landscape. The growing complexity of experiments, the necessity to quantify uncertainty, the pivotal role of hypothesis testing and inference, the incorporation of statistical analysis in experimental design and power analysis, and the emergence of advanced data analysis techniques have rendered statistics an indispensable tool in contemporary scientific research~\cite{Zdeborova2017, Mehta19}. During the last decades, statistics have empowered researchers to navigate complex data, derive valid conclusions, and make evidence-based decisions, playing an instrumental role in advancing scientific understanding~\cite{Barlow,Cousins, Vaux}.

In the last decade, machine learning (ML) methods have complemented the statistical analysis in Physics~\cite{Zdeborova2017, Carleo19, Guimera20, Agliari, Karagiorgi22}. ML has been used in processing satellite data in atmospheric physics~\cite{karpatne2018machine}, in weather forecasts~\cite{Jones17}, predicting the behaviour of systems of many particles~\cite{Carleo19}, discovering functional materials~\cite{Balachandran} and generating new organic molecules~\cite{Butler18}.  Indeed, recent works have shown that deep learning techniques outperform human-designed statistics~\cite{Haiman19}, providing evidence of the power of ML for analysing experimental data. Moreover, ML can discover new physical laws and equations. For example, symbolic regression~\cite{Udrescu20, Liu21} and sparse identification methods~\cite{Brunton16} have been used to derive physics equations from data.  Also, generative modelling offers a way to discern the most credible theory from various explanations for observational data. This is achieved solely through the data, without any predetermined understanding of the potential physical mechanisms operating within the studied system~\cite{schawinski2018exploring}. Therefore, the possibilities for using ML algorithms in physics range from experiments to theoretical analysis, opening up many opportunities. 

Although ML can help to address fundamental problems in physics~\cite{Karagiorgi22}, most physicists still do not recognise the importance of these methods and how they can help to discover functional patterns in data. ML has roots in Statistics and Computer Science~\cite{Bishop06, Murphy12, Goodfellow}, with applications from image segmentation to medical diagnosis~\cite{Murphy22}. Only recently, with new methods, such as deep learning~\cite{LeCun15, Goodfellow}, the area has increased its potential, allowing it to work with massive data. These improvements have made ML fundamental for physics discoveries, mainly where data is present~\cite{Zdeborova2017}. Also, many works have shown that Physics has the potential to develop new ML methods and help understand the ``black boxes'' algorithms, such as neural networks~\cite{Zdeborova2020}. Despite its potential, it is hard for physicists to grasp the immense literature available. Also, machine learning methods must be used carefully since wrong modelling choices or assumptions can result in misleading or unreliable conclusions~\cite{Mehta19}. Machine learning and statistical models should always be interpreted cautiously, considering the limitations of the data and potential pitfalls of the algorithms to avoid drawing erroneous or misleading inferences. For example, the p-value has been a subject of debate and criticism in scientific research due to several problems associated with its interpretation and use~\cite{Leek}. The primary objective of this review is to introduce machine learning methods to physicists, presenting both the fundamental concepts and concrete examples of their application.

This review delves into the fundamental concepts of ML, offers primary references, and outlines the steps to employ ML methods for physics discovery. We present the fundamental concepts, research papers, applications, and tools for utilizing ML in the field of physics while also exploring how physics can contribute to the development of ML.

\textbf{Basic concepts:} The main goal of ML is to find useful patterns in data~\cite{Bishop06, Murphy12}. In Physics, ML has been used when we have complex problems and lots of data~\cite{Mehta19}. For example, in the Large Hadron Collider (LHC) experiments at CERN, the collisions of protons or ions in several places around the circular collider generate tens of thousands of PetaBytes per year~\cite{Buckley}. However, since collisions are rare, it is necessary to classify the actual collisions as either interesting or uninteresting. Thus, ML methods filter the signal from the noise and help to search for new fundamental physics~\cite{Karagiorgi22}. 

ML can be divided into three main categories~\cite{Bishop06, Murphy12}, namely: (i) supervised learning, (ii) unsupervised learning and (iii) reinforcement learning. Supervised learning involves learning from labelled data to make predictions or classifications~\cite{Bishop06}, unsupervised learning discovers patterns and structures in unlabeled data~\cite{Murphy12}, and reinforcement learning focuses on an agent learning optimal behaviours through interactions with an environment and rewards~\cite{Sutton18}.

\begin{figure*}[!t]
\begin{center}
\includegraphics[width=1\linewidth]{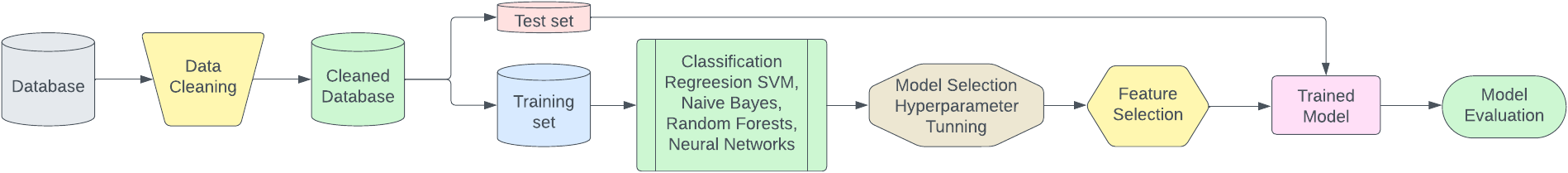}
\end{center}
\caption{The supervised learning pipeline. Typically, the data is divided into two sets. About 80\% of the data is allocated to the training set, where the model learns patterns and relationships from the data. The remaining 20\% forms the test set, which is an independent evaluation to gauge the model's performance and generalization abilities.}
\label{fig:ML}
\end{figure*}

\textbf{Supervised learning:} Supervised learning involves training a model based on a given set of examples~\cite{Murphy12, Mehta19}. Initially, we split the data set into two disjoint parts, one for training and another for testing. The training set is used to train a ML algorithm, which will learn the patterns in the data and then will be used to predict new observations on the test set. The data is composed of a set of examples, which are represented by the vectors $\{\mathbf{x}_i, y_i\}$, $i=1,2,\ldots, n$, where $\mathbf{x}_i$ is the set of attributes of the example $i$ and $y_i$ is the target variable we want to predict, which can be a single value or a vector. For example, to predict the ordered phase in the Ising model, $\mathbf{x}_i$, $i=1,2,\ldots, n$, can represent the spin configuration in a 2D lattice and $y_i$ is the phase, i.e., $y_i=1$ represents an ordered phase and $y_i=0$ a disordered one~\cite{Carrasquilla17}. When the target variable $y \in \mathbb{R}$, it indicates a regression problem. Conversely, if $y$ represents a class or label ($y \in \{C_1, C_2 \ldots, C_k\}$, where $C_j$ is a class), it becomes a classification problem. Predicting the ordered phase in the Ising model involves classification while estimating the time required to reach the ordered phase corresponds to a regression. Figure~\ref{fig:ML} summarizes the supervised learning process. 

ML algorithms aim to fit a model $f$ that maps the attributes to the target variable~\cite{Domingos12}. To elaborate, it is possible to fine-tune the function $f$ using input attributes denoted as $\mathbf{x}$ along with the target variable $y$. This fine-tuning process, performed during training, enables the derivation of the model's parameter set $\theta$, thus facilitating accurate predictions of $y$.
Mathematically,
\begin{equation}
y = f(\mathbf{x}, \theta) + \epsilon,
\end{equation}
where $\epsilon$ is the random error ($\langle \epsilon \rangle = 0$ and $\langle \epsilon_i\epsilon_j \rangle= 0$, $i,j=1,2,\ldots, n$, where $n$ is the number of observations in the training set). We can refer to $ f(\mathbf{x}, \theta)$ as the function that generates the data, and the goal of the ML algorithm is to estimate its set of parameters $\theta$ so that we can predict new observations as accurately as possible.

Generally, optimization methods are used to adjust a supervised learning model~\cite{Shrestha19}. In this case, we have to define a training loss function that will be minimized. This function can have different forms and depend on the type of learning process~\cite{Murphy22}. 
In general, the loss function $L$ compares the values of the target variable $y$ and the respective predictions $\hat{y}$. The average loss of the predictor on the training set is called the cost function, or empirical risk in decision theory~\cite{Murphy22},
\begin{equation}
\mathrm{C}(\hat{y}, y) = \frac{1}{n}\sum_{i=1}^n L(y_i, \hat{y}_i).
\end{equation}

In the case of classification, the choice of cost functions depends on the specific algorithm and objective~\cite{Rosasco04, Bishop06, Raschka18, Murphy22}. Typical examples include the 0-1 loss function and cross-entropy~\cite{Murphy22}. The 0-1 loss function measures the misclassification rate on the training set,
\begin{equation}
\mathrm{C}(\hat{y}, y) = \frac{1}{n}\sum_{i=1}^n I(\hat{y}_i \neq y_i), 
\end{equation}
where $I$ is the indicator function which returns one if and only if $y = \hat{y}$. 


In the case of regression, typical cost functions use the mean squared error loss~\cite{Murphy22}:
\begin{equation}\label{Eq:MSE}
\mathrm{MSE}(y, \hat{y}) = \frac{1}{n}\sum_{i=1}^{n}(y_i - \hat{y}_i)^2,
\end{equation}
or the mean absolute error loss~\cite{Murphy22}. Other functions are also possible (e.g.~\cite{Raschka18, Murphy22}). 

Thus, the model training process aims to discover a configuration of model parameters denoted as $\theta$, which serves to minimize the cost function when applied to the training dataset, i.e.,
\begin{equation}
\theta = \mathrm{argmin}_{
\theta} \frac{1}{n}\sum_{i=1}^n L(y_i, \hat{y}_i).
\end{equation}

Different algorithms can be applied to minimize the loss function, and their effectiveness depends on the problem and the algorithm's characteristics~\cite{Sun19}. Indeed, the ``No Free Lunch'' theorem states that, on average, no algorithm performs better than any other when considering all possible problems~\cite{Adam19}. It implies that the effectiveness of an algorithm is highly dependent on the specific characteristics of the problem at hand~\cite{Adam19}. These algorithms are based on different concepts~\cite{Murphy12}, such as probability theory (e.g. naive Bayes and logistic regression), decision trees, optimization techniques (e.g. support vector machines), neural networks and ensemble methods (e.g. random forests). All these methods can be used both for classification and regression. 

Naive Bayes is a probabilistic classification algorithm based on Bayes' theorem~\cite{Gareth13}. Given the class label, it assumes that the features are conditionally independent, hence the ``naive assumption''. 
It is computationally efficient and works well with high-dimensional data. On the other hand, logistic regression is a linear classification algorithm that models the relationship between the input variables and the probability of belonging to a specific class~\cite{Sperandei}. It is commonly used when the target variable is binary or categorical. Logistic regression can handle both numerical and categorical input features, and it provides interpretable coefficients that indicate the influence of each feature on the prediction~\cite{Murphy22}. 
Support Vector Machines (SVMs) are a powerful supervised learning algorithm for classification and regression tasks~\cite{Cortes, noble2006support}.
SVMs aim to find an optimal hyperplane that separates the data points of different classes with the largest margin~\cite{Cortes, noble2006support}. They can handle both linear and non-linear relationships using different kernel functions. Also based on optimization, neural networks, inspired by the structure and function of biological neural networks, are versatile models capable of learning complex patterns and relationships from data~\cite{Bishop95}. They consist of interconnected layers of artificial neurons that process information. Hebbian learning is used in neural networks to strengthen the connections between activated neurons, allowing the network to learn patterns from data~\cite{Alemanno23}. Deep neural networks with many hidden layers have led to breakthroughs in various domains, including computer vision, natural language processing, and speech recognition~\cite{LeCun15, Goodfellow}. 

Finally, random forests (RF) are an ensemble learning method that combines multiple decision trees to make predictions~\cite{Breiman}. Using a random subset of the data and features, RF  reduces overfitting, which occurs when a model becomes excessively fine-tuned to the training data, capturing noise and anomalies rather than the underlying patterns, leading to poor performance on unseen data. Random forests handle complex relationships, missing values, and identifying essential features, acting as a feature ranking algorithm~\cite{Murphy12}. 

We can employ various metrics to evaluate the performance of machine learning algorithms~\cite{Raschka18}. While accuracy, which returns  the fraction of correct predictions, is a commonly used measure for classification, caution is needed when dealing with unbalanced data as it can be misleading~\cite{wang2021review, kumar2021classification}. Additional vital metrics for binary classification include precision (proportion of true positive predictions out of all positive predictions), recall (proportion of true positive predictions out of all actual positive instances), F1 score (harmonic mean of precision and recall), and the area under the ROC (Receiver Operating Characteristic) curve~\cite{Raschka18}. In regression tasks, commonly used metrics include mean squared error (equation~(\ref{Eq:MSE})), mean absolute error, and the coefficient of determination ($R^2$)~\cite{Gareth13}, which measures the proportion of the variance in the target variable explained by the model. It is calculated as the square of the Pearson correlation coefficient between the actual and predicted values~\cite{Gareth13}.

Unlike traditional methods used in Physics to analyze data, ML does not use curve fitting~\cite{Gareth13}. Curve fitting aims to find the best fit for the existing data points, focusing on data approximation. However, minimizing the error in the training data is not the goal of ML algorithms. Training in machine learning refers to optimizing a model's parameters to predict unseen data by learning from labelled training examples. Curve fitting is related to overfitting~\cite{Bishop06}, which occurs when a model excessively performs well on the training data but fails to generalize to unseen data.
On the other hand, when the model is too simple and lacks the flexibility to capture the underlying patterns in the data, we have an underfitting~\cite{Bishop95}. The goal of machine learning algorithms is to find the balance between underfitting (high bias) and overfitting (high variance) in modelling data~\cite{Gareth13}. Bias refers to the error introduced by approximating a real-world problem with a simplified model. Variance, on the other hand, represents the variability of model predictions for different training datasets. The bias-variance tradeoff arises because reducing bias often increases variance, and vice versa~\cite{Gareth13}. Finding the optimal tradeoff is crucial for building models that generalize well to new data. The objective is to strike a balance where the model is complex enough to capture the underlying patterns but not overly complex to fit noise or irrelevant details.

There are many techniques to avoid overfitting~\cite{hawkins2004problem}. Cross-validation is a resampling technique that can decrease variance and assist in model selection and adjustment of hyperparameters~\cite{Raschka18}. Many ML models present several hyperparameters, which are values provided by the user. For instance, the number of layers in a neural network or the number of trees in the random forest algorithm are examples of hyperparameters. One common approach is the k-fold cross-validation. This method divides the dataset into k subsets or folds of approximately equal size. The model is then trained and evaluated k times, using a different fold as the validation set and the remaining folds as the training set. The performance metrics, such as accuracy or mean squared error, are recorded for each fold. The final performance measure is usually computed by averaging the results across the k iterations. Cross-validation helps decrease variance by providing a more robust estimate of the model's performance. It reduces the dependency on a single training-test split, which can be biased due to the randomness present in the data. By training and evaluating the model on multiple subsets of the data, cross-validation provides a more representative evaluation of the model's performance on unseen data.

Regularization techniques, such as L1 (Lasso) and L2 (Ridge) regression, can also avoid overfitting~\cite{Murphy22}, which adds a penalty term to the loss function during model training. This penalty discourages excessive complexity in the model by shrinking the magnitude of the coefficients. Regularization helps prevent overfitting by promoting simpler models that generalize well to unseen data. Dropout is a regularization technique commonly used in neural networks~\cite{Goodfellow}. It involves randomly disabling a fraction of the neurons during each training iteration. This helps prevent the network from relying too heavily on specific neurons or memorizing noise in the training data, reducing overfitting and promoting more robust generalization. 
Other methods to avoid overfitting include dropout in neural networks, early stopping, feature selection and dimensionality reduction, ensemble methods and data augmentation, which are techniques to artificially expand the training dataset by applying transformations, such as rotation, scaling, or flipping, to the existing data. By addressing overfitting, machine learning models can achieve better performance, improve robustness, and provide more accurate predictions in real-world applications. 

Supervised learning methods in Physics are mainly used in problems that involve classification, regression and time series forecasting~\cite{Carleo19}. For example, supervised learning algorithms have been used to classify particles in particle physics experiments, predict the Higgs boson's mass, and weather and climate forecasting~\cite{Carleo19, Mehta19, Karagiorgi22}.

\textbf{Unsupervised learning:} Unlike supervised learning, unsupervised learning is a type of machine learning where the algorithm learns patterns or structures from unlabeled data without any explicit target variable~\cite{Bishop06, Murphy12}. The goal is to discover hidden patterns, relationships, or clusters within the data. Examples of unsupervised learning methods include (i) clustering algorithms like k-means and hierarchical clustering; (ii) dimensionality reduction techniques, such as principal component analysis (PCA) and t-SNE; (iii) anomaly detection approaches, like one-class SVM and Isolation Forest; (iv) association rule learning, which are used to discover associations or patterns in large datasets, often applied to market basket analysis or recommendation systems; (v) generative models, like Gaussian Mixture Models (GMM) and Variational Autoencoders (VAEs), which learn the underlying data distribution and can generate new samples similar to the training data; (vi) topic modelling, which can be used to discover hidden topics or themes within a collection of documents; and (vii) density estimation, which can estimate the underlying probability distribution of a dataset. 

Unsupervised learning methods can be applied in various ways in physics. Principal component and clustering analysis can be used to identify phases and phase transitions of many-body systems~\cite{Wang}. 
Dimensionality reduction methods can assist in visualizing and understanding high-dimensional experimental or simulation data~\cite{Mehta19}. Anomaly detection algorithms can identify rare or unexpected events, outliers, or anomalies in experimental data, helping physicists to find potential errors, instrument malfunctions, or novel phenomena~\cite{Carleo19}. Generative models can generate synthetic data that matches the statistical properties of experimental observations, enabling model validation and exploration of new scenarios~\cite{schawinski2018exploring}.

Although with great potential to be applied in physics, data clustering presents several challenges~\cite{Rodriguez}. Firstly, the absence of ground truth labels makes evaluating clustering results subjective and dependent on heuristic measures. Secondly, determining the optimal number of clusters and selecting appropriate algorithms can be subjective and context-dependent~\cite{Rodriguez}. Additionally, the curse of dimensionality poses difficulties as higher-dimensional spaces make it harder to distinguish meaningful clusters~\cite{Rodriguez}. Complex and irregular data structures, scalability issues, and the interpretation of clustering results further contribute to the challenges. 

Overcoming these hurdles requires careful consideration of dataset characteristics, algorithm selection, parameter tuning, and the incorporation of domain knowledge~\cite{Rodriguez}. Ongoing research focuses on developing robust and scalable clustering methods to address the complexities of real-world data. The evaluation is also problematic since metrics like cluster purity, silhouette coefficient or adjusted rand index struggle to capture complex cluster structures, such as overlapping clusters, varying cluster densities, or non-spherical shapes~\cite{Murphy12}. Also, different clustering metrics may yield inconsistent results, making it challenging to compare and interpret the performance of different algorithms. Developing more robust and versatile clustering evaluation metrics is an ongoing research area, also for physicists~\cite{Zdeborova2020}.

\textbf{Semi-supervised learning (SSL):} Unlike traditional supervised learning, where a large labelled dataset is required for training, semi-supervised learning leverages a combination of labelled and unlabeled data. SSL can be a powerful tool for improving the accuracy of machine learning models. This is because the unlabeled data can be used to regularize the model, preventing overfitting. In Physics, SSL has been used, for example, to classify materials synthesis procedures~\cite{Huo2019} and detect distinct events in a large dataset of in tokamak discharges~\cite{Montes21}. The main algorithms for SSL~\cite{reddy2018semi} are (i) self-training, which can take any supervised method for classification or regression and modify it to work in a semi-supervised manner, taking advantage of labeled and unlabeled data; (ii) transductive SVM, which is a variation of support vector machines (SVMs) that is specifically designed for semi-supervised learning; (iii) label propagation, which assigns labels to unlabeled data by propagating labels from labeled data points to unlabeled data points that are similar to them; and (iv) ensemble methods, which combines multiple semi-supervised learning algorithms on different subsets of the data, and then their predictions are combined to make a final prediction.

\textbf{Reinforcement learning (RL):} In contrast to conventional machine learning techniques, reinforcement learning (RL) enables the extraction of knowledge from real-world experiences, surpassing the limitations of training data alone~\cite{Sutton18}. RL focuses on training agents to make sequential decisions in an interactive environment. Through trial and error, agents explore and exploit the environment, receiving rewards or penalties based on their actions. The goal is to learn an optimal policy that maximizes cumulative rewards. RL is used in scenarios without labelled training data and has applications in robotics, game-playing, and recommendation systems. For example, a RL system can master the game of chess solely from its rules, devoid of any preceding knowledge. By engaging in matches against adversaries or even self-play, the system progressively learns and hones its skills~\cite{Sutton18}. 

The main elements in RL are an agent and an environment it interacts with~\cite{reddy2018semi}. The environment provides the agent with information and feedback according to the agent's actions. The agent's primary goal is to maximize the obtained rewards the environment provides. Thus, the RL algorithms aim to learn an optimal policy maximising rewards. Given the observations, this policy defines the actions to take, thereby defining the agent's strategy. In this case, Markov Decision Process
is a mathematical framework used to solve these decision-making problems~\cite{Sutton18, dawid2022modern}. The main RL algorithms use Q-learning, which continuously learns
the optimal action-value function regardless of the policy followed during the training. This algorithm has many versions and can be implemented in neural networks~\cite{dawid2022modern}.
In physics, reinforcement learning is utilized for tasks such as control of quantum systems~\cite{fosel2018reinforcement},
create new experiments~\cite{melnikov2018active} , and discovering novel materials~\cite{moosavi2020role}.

\textbf{Deep Learning:} While classic supervised learning relies on manual feature engineering and simpler models, deep learning leverages deep neural networks to automatically learn representations from the data, making it capable of capturing complex patterns and achieving state-of-the-art performance in various domains~\cite{LeCun15, Goodfellow}. A neural network is a computational model composed of interconnected nodes, called neurons, organized into layers. Each neuron processes incoming information and produces an output, which becomes the input for other neurons in subsequent layers~\cite{Bishop95}. Through training, neural networks learn to adjust the weights of connections between neurons to recognize patterns and relationships in complex data effectively. This learning process enables them to make predictions, classify data, and solve various problems, such as image and speech recognition, natural language processing, and game playing~\cite{LeCun15}. 

With their ability to learn from examples and generalize from the data, neural networks have become a cornerstone of modern AI applications, driving remarkable advancements and innovations across numerous domains~\cite{LeCun15}. Many libraries are available to use in deep learning, including (i) TensorFlow~\cite{Pang2020}, which is an open-source software library powered by Google Brain, (ii) PyTorch~\cite{Paszke}, which Facebook originally developed; (iii) Scikit-learn~\cite{Scikit-learn}, which provides a wide range of algorithms for supervised, unsupervised, and reinforcement learning. Deep learning methods can be used for all the tasks discussed previously. 

Deep neural networks can also predict time series~\cite{lim2021time}. The most popular approaches include (i) recurrent neural networks, (ii) convolutional neural networks and (iii) transformers. Mainly, transformers, although developed for natural language processing tasks, treat the time series data as a sequence of words or characters. These models have been used to predict dynamical systems representative of physical phenomena~\cite{geneva2022transformers}.

\textbf{Physics informed machine learning:} Deep neural networks have also been used to solve partial differential equations, which enable scientific prediction and discovery from incomplete models and incomplete data~\cite{raissi2019physics}. In this approach, called physics-informed machine learning, models are trained on both data and physical principles. More specifically, the cost function is changed to include a term that penalizes the model for violating the physical principles~\cite{karniadakis2021physics}. For example, if we are trying to model the behaviour of a fluid, we might add the Navier-Stokes equations as a constraint to the cost function. This ensures that the machine learning model is consistent with the known laws of physics, which can help to improve the accuracy of the model. Therefore, this approach combines deep learning with prior knowledge about the fundamental laws and principles of physics to create more accurate and reliable models of the physical world~\cite{raissi2019physics, karniadakis2021physics}.

\textbf{Physical discovery:} Machine learning methods have also been used for physics discovery. Mainly, 
symbolic regression searches the space of mathematical expressions to find the model that best fits a given dataset~\cite{Brunton16, udrescu2020ai}. No particular model is provided as a starting point for symbolic regression. For example, when applied to 100 equations from the Feynman Lectures on Physics, symbolic regression discovered all of them~\cite{udrescu2020ai}. Most methods for symbolic regression are based on genetic algorithms~\cite{Brunton16}.

\textbf{Causal inference:} Most machine learning methods do not consider the causal relationships between variables. Only recently, causal machine learning methods have been designed to identify the causal relationships between variables and to use this information to make better predictions~ (e.g.~\cite{runge2015identifying}). Causal inference methods can make causal claims about the world, even with confounding variables. In physics, causality can be used to infer the connection between variables. For instance, in complex systems, causality methods have been used to infer the structure of the underlying system, like in the brain~\cite{alves2023diagnosis} and climate systems~\cite{runge2015identifying}.

\textbf{Perspectives} While ML has demonstrated successful applications in the realm of Physics, several challenges and adjustments still must be addressed~\cite{Carleo19, Mehta19, dawid2022modern}. A recurring hurdle is the scarcity of data suitable for training ML models, often accompanied by the predicament of imbalanced data—where significant class imparities exist, hampering model accuracy. Furthermore, scalability emerges as a concern, given the computationally intensive nature of training and deploying ML models, particularly for vast datasets~\cite{Mehta19}. The inherent opacity of ML model decision-making poses an additional obstacle in comprehending their predictions~\cite{Zdeborova2020}. In these scenarios, physicists can rise to the occasion, devising strategies to surmount these obstacles and tailoring approaches for enhanced knowledge extraction within physical systems~\cite{Zdeborova2020}. Particularly, statistical physics can shed some light on these problems, contributing to the development of ML~\cite{Zdeborova2020, Agliari20}.

\textbf{Additional material:} We made available a Jupyter notebook to put in practice most of the concepts presented here, mainly associated to supervised and unsupervised learning. The code can be used as a starting guide for analysing data in Python. The reader can access the code in the following link:\\ \url{https://github.com/franciscorodrigues-usp/MLP}

\acknowledgments
Francisco Rodrigues acknowledges CNPq (grant 309266/2019-0) for the financial support given for this research. 

\bibliographystyle{unsrt}
\bibliography{references}

\end{document}